\documentclass{Interspeech2024}

\interspeechcameraready

\title{Positional Description for Numerical Normalization}

\name[]{Deepanshu}{Gupta}
\name[]{Javier}{Latorre}

\address{
   Apple
   }
\email{dkg@apple.com, jlatorremartinez@apple.com}

\begin{document}

\maketitle

\begin{abstract}

We present a Positional Description Scheme (PDS) tailored for digit sequences, integrating placeholder value information for each digit. 
Given the structural limitations of subword tokenization algorithms, language models encounter critical Text Normalization (TN) challenges \cite{sproat-2022-boring} when handling numerical tasks. 
Our schema addresses this challenge through straightforward pre-processing, preserving the model architecture while significantly simplifying number normalization, rendering the problem tractable. 
This simplifies the task and facilitates more compact production-ready models capable of learning from smaller datasets. 
Furthermore, our investigations reveal that PDS enhances the arithmetic processing capabilities of language models, resulting in a relative accuracy improvement of 23\% to 51\% on complex arithmetic tasks. 
We demonstrate that PDS effectively mitigates fatal numerical normalization errors in neural models, requiring only a modest amount of training data without rule-based Finite State Transducers (FST). 
We demonstrate that PDS is essential for both the Text-To-Speech and Speech Recognition text processing, enabling effective TN under production constraints.

\end{abstract}
\section{Introduction}


Text Normalization (TN) is  the first step in a Text-to-Speech (TTS) pipeline, converting  non-standard words such as  digit or acronyms into words \cite{sproat2001normalization}. 
For TN, more often than not there is one single acceptable output and the errors are immediately noticeable. 
This essential pre-processing step is part of ensuring that the synthesized speech is fluent, natural, and conveys the intended meaning. 
Traditionally, TN has used a rules systems such as finite-state transducers (FST)\cite{ebden2015kestrel, ritchie-etal-2020-data}. 
This is due to 2 reasons -- (i) need for high precision and (ii) faster processing.

Neural sequence-to-sequence (NS2S) models\cite{sproat2016rnn, zhang2019neural}, often fall short when confronted with numerical expressions, particularly in contexts where the place-value system of numbers plays a pivotal role\cite{sproat2016rnn, adc2020}.

In NS2S, the first step is usually an automatic subword level tokenization of both the input and output\cite{Mansfield2019}. 
Then, the model learns the mapping between input and output tokens. 
A core issue of NS2S models for TN is their limited ability to model the place-value system in numbers, which is a by-product of tokenization scheme used in these end to end models. 
The standard tokenization treat numbers as any other sequences of subword tokens, leaving the model to decipher the appropriate place value for each digit. 
Since it's almost impossible to cover all the different digit and placeholder combinations this leads to severe data scarcity, even with massive datasets. 
The subword tokenization schemes directly used by today's language models including BPE\cite{sennrich-etal-2016-neural}, Sentencepiece\cite{kudo-richardson-2018-sentencepiece} etc. introduce implicit bias in the model, towards 2-3 digit sub-string seen in the training data, forcing the models to learn sparse mappings. 
This implicit bias leads to intractable and almost infinite number of input-output mappings, which in turn leads model to produce ``hallucinations," where the model incorrectly assigns place values, resulting in inaccurate and contextually inappropriate normalizations, called "fatal errors" \cite{sproat2016rnn, sproat-2022-boring}.  
For example: a subtoken ``@@23" in string ``1234 apples." might be mapped to an entirely different normalization than a ``123 apples". 
Thus the model has learn context, normalization and number-string structure to learn normalization correctly. 
This leads to the impression that number normalization is non-learnable challenge using NS2S models, even though they could potentially be learning the right context.

While for some languages like English a relatively simple set of rules can cover most of the cases, in others such as Russian/Lithuanian/Polish etc. numbers are \textit{declined} so they have to be normalized differently depending on the context. 
Similarly, in code mixing sentences any digit is an inter-lingual homograph disambiguation which might depend on extended contexts that are difficult to handle via rules.

In data rich scenario like English, we notice that fatal numerical normalization is much more prevalent in large numbers. 
Our exploratory analysis found that TN models trained without PDS are only 40\% accurate on numbers above a million ($10^6$), while their output is only 10\% accurate when the number is greater than a billion($10^9$). 
This analysis was done on 1000 examples test set generated synthetically.

Advances in Large Language Models (LLM) might tempt one to claim this is a problem that can be solved with more data or larger models. 
For example, different versions of ChatGPT \cite{zhang2023chat} were evaluated with some success on the English Text-Normalization challenge.
Though, LLMs have a great understanding of the context, they still suffer from the curse of tokenization due to subword algorithms.

Our contributions are as follows: 
\\(i) PDS makes numerical normalization input-output pairs tractable, which in turn vastly reduces fatal errors  
\\(ii) PDS is applicable without any changes required in model architecture, leads to learning similar or better behavior with less data and/or fewer model parameters and/or less training time
\\(iii) Finally, we demonstrate that PDS can be used in language agnostic manner to develop a reliable normalization system using small scale data collection 

Note, PDS doesn't make a model learn the context better, but only resolves the limitations introduced by tokenization algorithms to learn numerical normalization. 
\section{Related work}

\subsection{Tokenization Scheme}
Our scheme closely aligns with San-Segundo et al.'s trainable Multi-Lingual TN system \cite{sansegundo13_ssw}. 
They use a method where numbers are represented as a \textit{digit-underscore-placevalue} approach, which is very similar to how we represent digits, with few differences in format and approach. 
We similarly frame TN as a Machine Translation problem, but instead of Phrase Statistical Machine Translation systems, we use Transformer based Machine Translation models. 
One clear differnce of focus between our work and theirs is the emphasis of our work specifically on fatal errors and reducing them to almost negligible amount with PDS.

\subsection{FST Systems} 
In `Kestrel Text Processing System" \cite{sproat2001normalization}, the TN problem was broken into various semiotic classes, laying the groundwork for understanding different types of normalization challenges and their interpretation. 
Kestrel introduces a
FST-based solution,
providing insights into practical approaches for linguistic processing.
Kestrel system is used to generate the data \cite{sproat2016rnn} that we use to demonstrate the effectiveness of our approach. 

\subsection{Neural+FST Systems} 
In \cite{sproat2016rnn} a substantial TN dataset encompassing English, Polish, and Russian is presented. 
This dataset has become a standard benchmark for evaluation in the field. 
They highlight the limitations of RNN/LSTM-based translation models, such as erroneously normalizing ``300" as ``four hundred." 
To address these challenges, they propose a multi-stage normalization system, leveraging methods like taxonomy prediction to enhance the accuracy of predictions.
Similarly, Proteno\cite{tyagi2021proteno}, Transformer based Text Normalization\cite{ro2022transformerbased}, and Unified Text Normalization\cite{lai2021unified} advocate for using tagger-based semiotic class identification followed by FST-based verbalization strategies to enhance the TN processes.

An  extensive analysis on the challenging nature of TN, characterizing the encountered errors as "unrecoverable" can be found in  \cite{zhang2019neural} and \cite{Roark2019}.
These two papers explore various neural methods, including Transformer architectures and hybrid models, such as a sequence of semiotic class identification followed by sequence-to-sequence verbalization. 
Another hybrid approach involves using a sequence-to-sequence model for decoding, incorporating FST-driven verbalization techniques. Notably, they demonstrate the ability to learn grammars from smaller datasets. 
In \cite{Sunkara20201} and \cite{pusateri2017mostly} inverse text normalization(ITN) was addressed in a similar fashion.

These TN systems employ a 2 stage pipeline to achieve better reliability for fatal errors, mostly concerned with accurate number normalization. 
Our experiments show that we can achieve similar resiliency using PSD and keeping the model simple. 

\subsection{FST + Language Model}
Recently, NeMO \cite{zhang2021nemo} propose a unique, data-efficient method for TN, treating it as a MASK/PREDICT problem where a span is masked, and the most probable verbalization is selected based on language modeling metrics. 
They confine the potential masked predictions to outputs generated by language-specific FSTs, offering a distinct approach to the normalization task.
This late stage decision making makes it interesting and different from previous Hybrid FST + ML based solutions. 
This utilizes the language model capabilities of the recent Large Language Models, so theoretically it doesn't need any human annotated dataset. 
However, this late stage decision making comes at the cost of higher inference time, since all the various combinations of normalizations FST need to be done and then run through a ranking model.
Such a system is very useful as an offline data generation tool.

\subsection{End to End Neural Methods}
Other works model TN as a Translation task from a source sentence to a target sentence, without any rules/heuristics in between. 
In \cite{Mansfield2019}  subword-tokenizations was used to reduce the size of vocabulary and do TN on sentence level instead of word level. 
These authors do not employ any pre-processing or post-processing, hence resulting in many numerical fatal errors. 
Other related works such as \cite{adc2020} have gone further treating TN and Grapheme to Phoeneme as a single challenge of training text processing, using Tranformer encoder-decoder models to train on both task in a single model. 
In Memory Augmented text normalization\cite{pramanik2019text} the authors propose architectural changes and more intricate model engineering techniques to mitigate the occurrence of fatal errors in TN.
Similarly, we also treat TN as a translation task on sentence level, without any hybrid class prediction or modification rules. 

\subsection{LLM based methods}
Recently, Zhang et al. \cite{zhang2023chat} discussed and evaluated ChatGPT on the task of TN using the data from \cite{sproat2016rnn}. 
They demonstrate that ChatGPT is better capable of understanding the right context of a sentence, even identifying the incorrect normalization in the original datasets. 
While agreeing with their main observation, we have also observed that  ChatGPT too suffers from fatal numerical normalization when numbers are larger than a million. 
When we evaluate ChatGPT for Non-English TN, the performance is significantly worse. 
One example is the number ``28" which seems to incorrectly normalized by ChatGPT 3.5 for 23 languages out of the 67 we evaluated. 
Additionally, for practical TTS applications, the TN needs to be completed in single digit millisecond latency.
So using a billion scale parameter inference is impractically expensive and/or too slow.

\section{Methodology}
In this section, we describe how PDS works, and how it theoretically helps make the problem of TN a more tractable problem.

PDS is extremely simple. 
It is inspired by how explicit characters are used in Chinese/Japanese to describe the placevalue information of a number. 
In Chinese ``123" is written as ``1 hundred 2 ten 3", which can be used similarly as additional signal to do number normalization correctly. 
For example, ``I have 123 apples" becomes ``I have \_ 1 03 2 02 3 01 \_ apples". 
We find it is very similar to how \cite{sansegundo13_ssw} tokenize numbers. 
\footnote{
They write ``123" as ``1\_3 2\_2 3\_1", however they introduce some minor redundancy by having each placevalue and facevalue combination as a single token, where our tokenization scheme can reuse the placevalue symbols. 
}
In this way, every number in the input string is ``described"  by  rewriting it in 2 parts: a single digit token with the face-value and a double digit token  with its placevalue. 
Additionally, we use ``\_" to indicate the digit boundary. 

A simple algorithmic way to represent this is : 
\begin{python}
def apply_pds(number = "123"):
    for idx, digit in enumerate(number):
        facevalue = digit 
        placevalue = len(number) - idx 
        yield facevalue, placevalue
\end{python}

This allows the models to understand presence of ``03" means ``hundreds" or ``12" means ``hundred billion" etc. 
In an extremely simplistic scenario, with max 20 digits normalization, this leads up to $10$ * $20$, or 200 different combinations.\footnote{number of digits [0-9] * maximum number of digits for normalization, (in our case hundred quintillion or $10^{20}$)}
This makes the input-output objective a very simple one-to-one map. 
Theoretically, this means as low as 200 examples will be enough to learn number expansions up to $10^{20}$. 
This makes the input-output mapping of the data more straightforward for the model to learn, and with fewer examples. 
So, this helps train higher quality models when human annotated data is expensive to create.

PDS is very flexible, and can be made more complex in turn to simplify the numerical normalization process.
With algorithm shown, we assume absence of any other characters in the number. 
To extend it we can have thousands separator like ``," in US-English or ``." or whitespace in other European languages. 
PDS can be modified to take those separators into account, and remove the decimal separators.
For example: ``1,234" becomes ``\_ 1 04 2 03 3 02 4 01"
However, in favor of minimal preprocessing, we do not do any such removal. 

\section{Experiments}

\subsection{Data \& Pre-trained Models}
To evaluate the effectiveness of PDS in TN we use the TN data generated by Google's Kestrel Text Normalization System\cite{sproat2001normalization} in English, Polish and Russian. 
From that data, first we separated for each language a uniformly sampled held-out 10k test set.
Additionally, to test each semiotic class independently we create heldout test sets of 1k examples per class. 

For training, we sample 3 different data sizes 100k, 1M and 10M sentences respectively.
Over these sets we fine tuned the Google mt5-small model \cite{xue2021mt5} with and without applying PDS over the training and test data. This yields a total of 9 models (3 languages x 3 set-sizes).  
Each of these 9 models was trained for different number of steps(10k, 25k, 50k, 100k) and we measure the performance over the test set using Exact Match Accuracy which calculates exact match at sentence level. 
We train all our models with a batch size of 32, on a single NVIDIA A100 GPUs machines. 

\subsection{Results}
\begin{figure}[!htbp]
   \centering
   \includegraphics[width=0.47\textwidth]{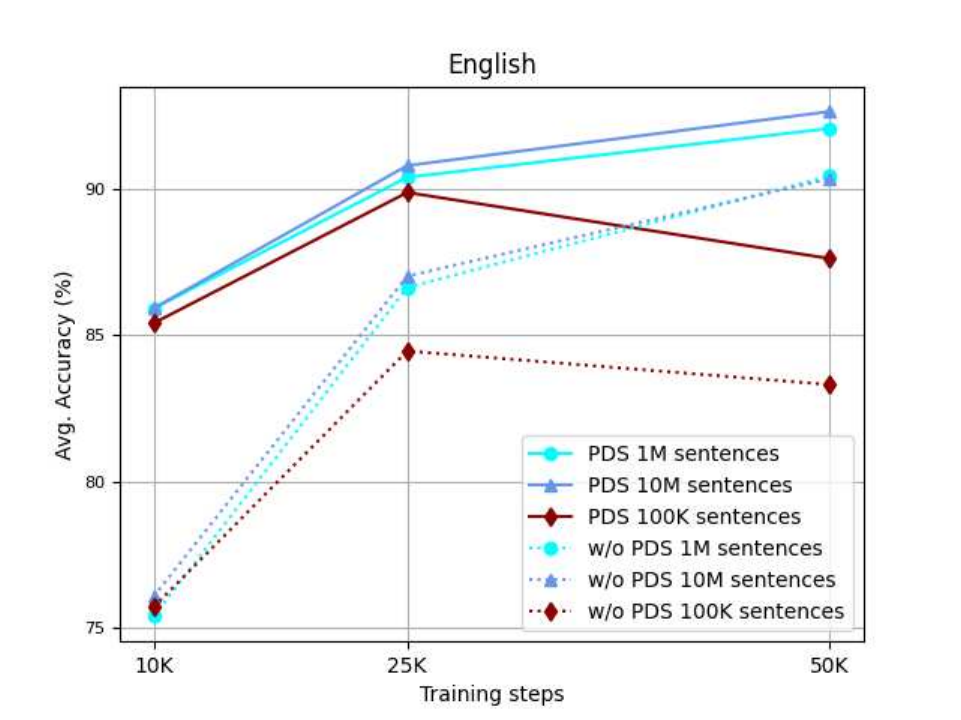}
   \caption{Avg. accuracy for English for different data sizes.}
   \label{fig:TN_English}
\end{figure}

\begin{figure}[!htbp]
   \centering
   \includegraphics[width=0.47\textwidth]{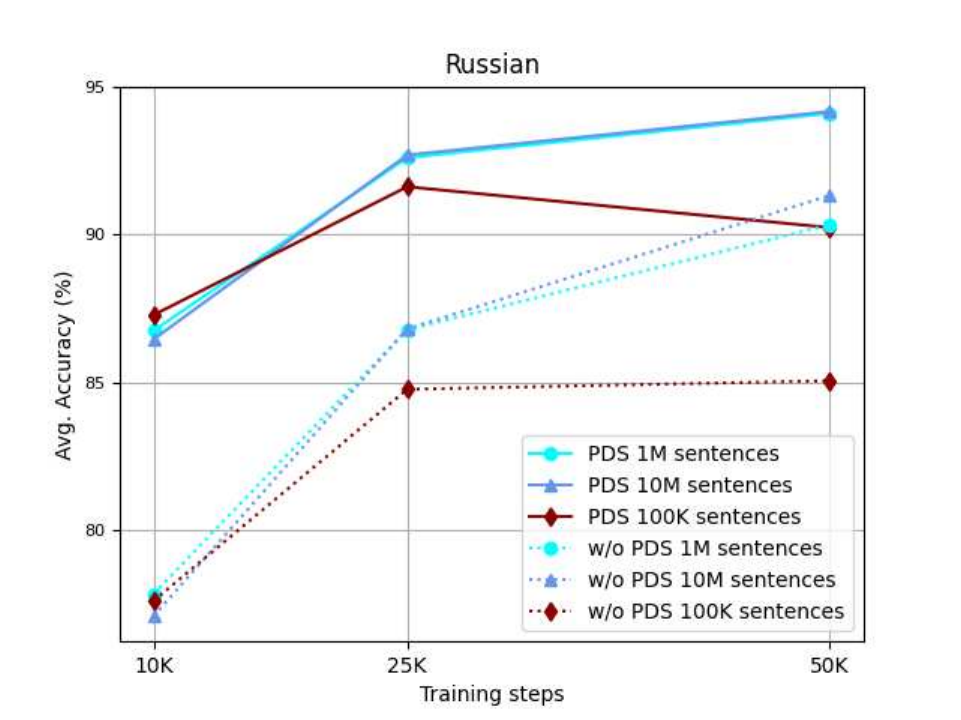}
   \caption{Avg. accuracy for Russian for different data sizes.}
   \label{fig:TN_Russian}
\end{figure}

\begin{figure}[!htbp]
   \centering
     \includegraphics[width=0.47\textwidth]{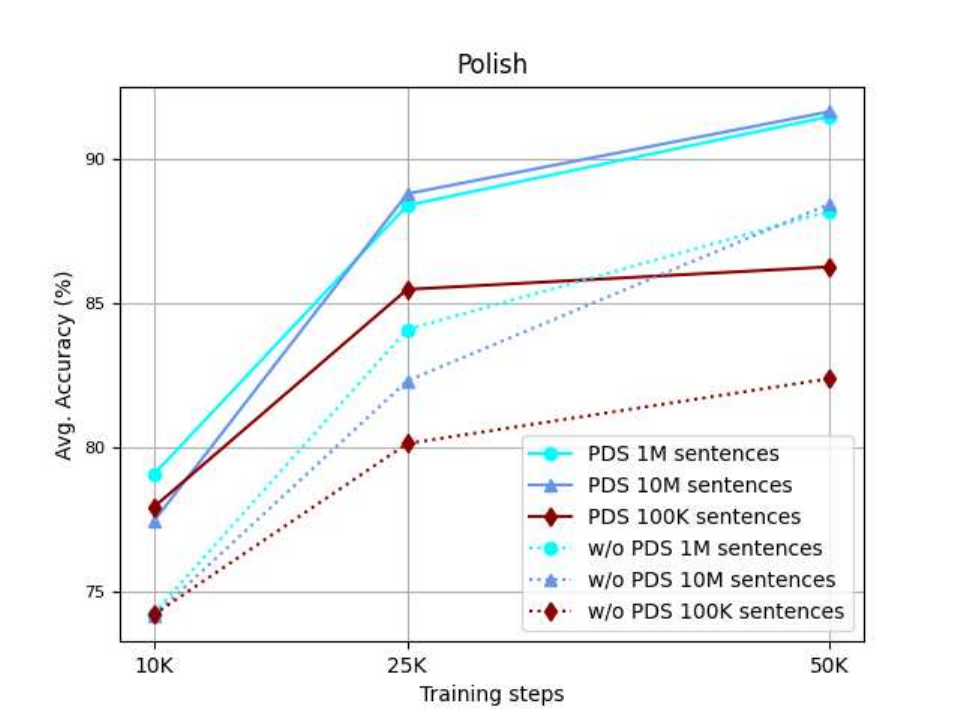}
   \caption{Avg. accuracy for Polish for different data sizes.}
   \label{fig:TN_Polish}
\end{figure}

Figures \ref{fig:TN_English}, \ref{fig:TN_Russian} and \ref{fig:TN_Polish} show the average TN accuracy across the different categories \footnote{Results for ELECTRONIC, ADDRESS, FRACTION, and TELEPHONE were removed from the average because their datasets are highly unreliable} for English, Russian and Polish, for models with and w/o PDS trained on different amounts of data and for different number of steps. 
The evaluation was computed on uniformly sampled testset with 10k examples that contain more than half of the examples where no normalization is required. 
These redundant examples are needed to test the model's sanity so model doesn't change every example given to it. 

For the same amount of fine-tuning data and training steps, the models using PDS are consistently better. Actually, models fine-tuned w/o PDS only become better than those using PDS when they trained in at least 10 times more data and/or for double the number of steps.

\begin{table*}[t]
    \centering
    \begin{tabular}{|c|c|c|c|c|c|c|}
        \hline
        &\multicolumn{2}{|c|}{English}&\multicolumn{2}{|c|}{Polish}&\multicolumn{2}{|c|}{Russian}\\
        Category & w/o PDS & PDS & w/o PDS & PDS & w/o PDS & PDS\\
        \hline
        MEASURE & 91.80\% & \textbf{94.20\%} &  80.70\% & \textbf{84.80\%} & 85.20\% & \textbf{86.30\%}\\
        TIME & 49.10\% & \textbf{79.70\%} & 28.06\% & \textbf{57.55\%}& 32.07\% & \textbf{75.51\%}\\
        LETTERS & 88.20\% & \textbf{88.40\%} & 89.20\% & \textbf{90.40\%} & 94.40\% & \textbf{94.90\%}\\
        ORDINAL & 95.30\% & \textbf{95.40\%} & \textbf{91.00\%} & 90.30\% & 95.50\% & \textbf{95.90\%}\\
        DIGIT & 44.90\% & \textbf{54.20\%} & 45.98\% & \textbf{64.37\%} & 80.62\% & \textbf{93.02\%}\\
        DATE & 99.10\% & \textbf{99.50\%} & \textbf{96.80\%} & 95.80\% & 95.20\% & 95.20\%\\
        VERBATIM & 94.40\% & \textbf{94.50\%} & 97.90\% & 97.90\% & \textbf{98.60\%} & 98.70\%\\
        PLAIN & \textbf{99.30\%} & 99.20\% & 99.80\% & 99.80\% & 99.80\% & 99.80\%\\
        CARDINAL & 95.60\% & \textbf{96.90\%}  & 88.80\% & \textbf{90.90\%} & 88.80\% & \textbf{90.90\%}\\
        DECIMAL & 86.90\% & \textbf{96.80\%} & 77.50\% & \textbf{85.90\%} & 77.50\% & \textbf{85.90\%} \\
        Macro Avg. Accuracy & 84.46\% & \textbf{89.88\%} & 84.77\% & \textbf{91.61\%} & 84.77\% & \textbf{91.61\%} \\
        \hline
    \end{tabular}
    \caption{Comparison of Accuracy with and without PDS for model trained on 100k dataset for 25k timesteps}
    \label{tab:pds_accuracy_comparison}
\end{table*}

Table \ref{tab:pds_accuracy_comparison} presents a more dissected view of the accuracy across  all different semiotic classes.
We refer to \cite{sproat2001normalization} for a description of each semiotic class.  
For all categories where there is no numerical normalization, the models perform almost the same. 
However, in almost all categories with numerical normalization PDS helps make models better. 
While the metrics may look insignificant in certain categories, it is important to look at the errors each model makes to make a fair assessment of the utility of the PDS.
It is those few errors that eventually make up the ``fatal errors".
We refer to Section \ref{discussion} for more closer details into the errors.

\subsection{Experiments on Addition Multiplication Subtraction task}
Inspired by \cite{shen2023positional} we also conducted a set of simple experiments to understand if PDS can also help models learn arithmetic operations. 
To test this we create a 10k and 25k synthetic dataset of 2-to-5 operation equations involving a mix of addition, subtraction and multiplication similar to BODMAS style equations.
An example of the data looks like 
``$377 * 11 - 776 + 765 = 4136$"
We generate random numbers between $0$ to $10^{10}$. 
We train these models for 100k steps and evaluate on a synthetically generated held out set of 1000 arithmetic equations. 

\begin{table}[htbp]
    \label{tab:arithmatic}
    \centering
    \begin{tabular}{|c|c|c|}
        \hline
        \textbf{Data Size} & \textbf{w/o PDS} &  \textbf{PDS} \\ \hline
        10k       & 0.38519 &\textbf{0.47704}           \\ \hline
        25k           & 0.39852        & \textbf{0.59333}           \\ \hline
    \end{tabular}

    \caption{Arithmetic Task Average Accuracy}
\end{table}

\section{Discussion}\label{discussion}

We examine the different kind of errors made by each model, their severity and impact on customer. 
Firstly, we categorize up to 20 randomly selected errors from each class in following internally defined buckets:
\\
(i) IGNORABLE or incorrect class but correct normalization \\
(ii) CRITICAL or correct class but incorrect normalization \\
(iii) FATAL or incorrect class and incorrect normalization 

For our English model finetuned with PDS on 100k examples and 25k time steps, we find $0$ CRITICAL/FATAL category errors, which is the primary motivation of PDS. 
On the other hand we found $13 / 200$ examples in model trained without PDS to be either CRITICAL/FATAL. 

A couple of examples, where models w/o PDS made CRITICAL/FATAL mistakes, but models with PDS didn't are presented as following:

\textbf{Example 1:} FATAL \\
\textsc{Original Example}  There is no access from the 6318  Military Road.\\
\textsc{Output (w/o PDS)} There is no access from the six hundred thirty one thousand eight hundred eighteen
Military Road.\\
\textsc{Output (w/ PDS)}  There is no access from the six three one eight Military Road.

\textbf{Example 2:} CRITICAL \\
\textsc{Original Example} Event occurs at 23:54.\\
\textsc{Output (w/o PDS)} Event occurs at two forty eight seconds\\
\textsc{Output (w/ PDS)} Event occurs at twenty three fifty four

\textbf{Example 3:} CRITICAL \\
\textsc{Original Example} Biffle remained quickest with a time of 49.297 seconds.\\
\textsc{Output (w/o PDS)} Biffle remained quickest with a time of forty nine
 point two seven seven seconds\\
\textsc{Output (w/ PDS)}  Biffle remained quickest with a time of forty nine point two nine seven seconds

Additionally, our results are verified by human listeners internally on production sized small bilingual English-Spanish models with 70\% relative error reduction when using PDS vs without using PDS. 
In internal ablation, we also found that models w/o PDS accuracy on cardinal number normalization for numbers above a billion is 11\%, while PDS models achieve a 98\% accuracy on a small 1000 random example test set, trained using just 10k examples. 

\section{Conclusion}

In this work, we demonstrate that very simple pre-processing can allow us to achieve computational \& data efficiency and accuracy for critical consumer facing applications. 
The usage of a simple Positional Description scheme (PDS) almost halves the number of errors in numerical TN in many cases, eliminates fatal errors completely, and allows for reliable production size models.

We believe using PDS is a promising approach to mitigate fatal number normalization errors, especially in those applications where rules are not an option.
Experimental results on our internal dataset validate this hypothesis for code-switching texts in languages written in the same script. 


\bibliographystyle{unsrt}
\bibliography{interspeech2024}

\end{document}